\pdfoutput=1
\documentclass[11pt]{article}

\usepackage{ACL2023}

\usepackage{times}
\usepackage{latexsym}

\usepackage[T1]{fontenc}
\usepackage[utf8]{inputenc}

\usepackage{microtype}

\usepackage{inconsolata}
\usepackage{balance}
\usepackage{todonotes}
\usepackage{cleveref}
\title{Putting Natural in Natural Language Processing}
\author{Grzegorz Chrupała\\
  Department of Cognitive Science and Artificial Intelligence\\
  Tilburg University\\
  \texttt{grzegorz@chrupala.me}}

\begin{document}
\maketitle
\begin{abstract}
Human language is firstly spoken and only secondarily written. Text, however, is a very convenient and efficient representation of
language, and modern civilization has made it ubiquitous. Thus the field of NLP has overwhelmingly focused on processing written rather
than spoken language. Work on spoken language, on the other hand, has been siloed off within the largely separate speech processing
community which has been inordinately preoccupied with transcribing speech into text. Recent advances in deep learning have led to a
fortuitous convergence in methods between speech processing and mainstream NLP. Arguably, the time is ripe for a unification of these
two fields, and for starting to take spoken language seriously as the primary mode of human communication. Truly natural language processing
could lead to better integration with the rest of language science and could lead to systems which are more data-efficient and more
human-like, and which can communicate beyond the textual modality.

\end{abstract}

\section{Introduction}
\label{sec:intro}
The ACL 2023 theme track urges the community to check the reality of the
progress in NLP. This position paper adopts an expansive interpretation
of this question. It is definitely worth inquiring into the apparent
advances of current NLP in their own terms. Here, however, I question
these terms and argue that our field has focused on only a 
limited subset of human language which happens to be convenient to
work with, and thus misses major aspects of human communication.

\subsection{Human Language is Primarily Spoken} 
\label{sec:lang-spoken}

Humans are an exceptional species in many ways, and out of these, human
language is one of the most salient. Unlike communication systems
used by other organisms, human language is open-ended, capable of
expressing abstract concepts, and of reference to events
displaced in time and space. While the capacity to acquire language 
is universal and largely innate \citep{descent_of_man, pinker_bloom_1990}
it also is culturally mediated and likely arose via gene-culture
co-evolution \citep{deacon1998symbolic,richerson2010possibly}.

One revolutionary technology which turbo-charged human
language was writing, which was invented a handful of times in the most
recent few thousand years of the human story
\citep{fischer2003history}.
Writing, followed by the
printing press, followed by the Internet, have made written text
ubiquitous to the extent that it is easy to forget that the primary
and universal modality for most human communication throughout history
has been spoken.\footnote{I am using \emph{spoken language} in the
  broad sense here, including both the oral and gestural (signed)
  modes of expression, and opposing these to the written modality.}

Even today many of the world's languages do not have a standardized
written form. For those that do, the written modality
 originated as a compressed, symbolic representation of the spoken
 form.

Children acquire a spoken language (and not infrequently two or more)
within the first few years of their life with no or little explicit
instruction, largely relying on weak, noisy supervision via social
interaction and perceptual grounding. In contrast, they require
hundreds of hours of explicit instruction and arduous conscious
practice to learn to read and write, and most are only able to learn
the written modality a couple of years at best after becoming fluent
communicators in one or more spoken languages.

\subsection{Reality check}
Thus, arguably, the natural language for which we are biologically
equipped is spoken. Written language is a secondary development, which
happens to be very useful and widespread, but is nevertheless
derivative of speech. This appears to be the consensus view in
linguistics going back at least a century
\citep{desaussure1916,bloomfield1933}.\footnote{However see
  \citet{doi:10.1080/02702710600846803} for a dissenting view.}

Given these facts, is then the field of Natural Language Processing
(NLP) a misnomer? Are we making less progress with getting machines to
communicate via human language than current advances with processing
written text would have us believe?

\section{NLP is Written Language Processing}
\label{sec:nlp-written}
%The field of Natural Language Processing is represented at the main
%conferences (ACL, AACL, NAACL, EACL, EMNLP) and journals (TACL, CL) run by
%the Association for Computational Linguistics (ACL), and a few other
%similar conferences such as COLING and LREC.
To anyone with 
experience reading, reviewing and publishing papers in NLP conferences
and journals (such the ACL conferences and TACL) it
is evident that the field is very strongly focused on processing
written language. While this is evident to practitioners, it is also
largely tacit and implicit.
%to the point that it might seem superfluous to even
%inquire into this fact.

\subsection{Unstated assumptions}
\label{sec:unstated}
The fact that a paper is concerned with written as opposed to spoken
oral or sign language is almost invariably assumed to be the default
and not explicitly stated.
%-- similarly to how many papers do not
%explicitly state that they focus on English.
Furthermore, even if
there is some interest in tackling a dataset of originally spoken
language (for example in much work on dialog and child language
acquisition), the usual approach is to use a written transcription of
this data rather than the actual audio. This is partly a
matter of convenience, but partly due to the assumption
that the written form of language is the canonical one while the
audio modality is just a weird, cumbersome encoding of it.

To some extent such an implicit belief also lurks in much work within
the speech community: the main thrust of speech research has always
been on so called Automatic Speech Recognition (ASR), by which is meant
automatically transcribing spoken language into a written
form. Written text is treated as an interface and an
abstraction barrier between the field of speech processing and NLP. In
\Cref{sec:harder,sec:unify} I address problems arising from
the above assumptions, as well as the challenges and opportunities we
have once we discard them. Firstly, however, it will be instructive to briefly quantify the
assertion that NLP is Written Language Processing.
by looking at historical publication patterns.

\subsection{Publication patterns}
\label{sec:pubpattern}

\begin{figure}
  \centering
  \includegraphics[width=\columnwidth]{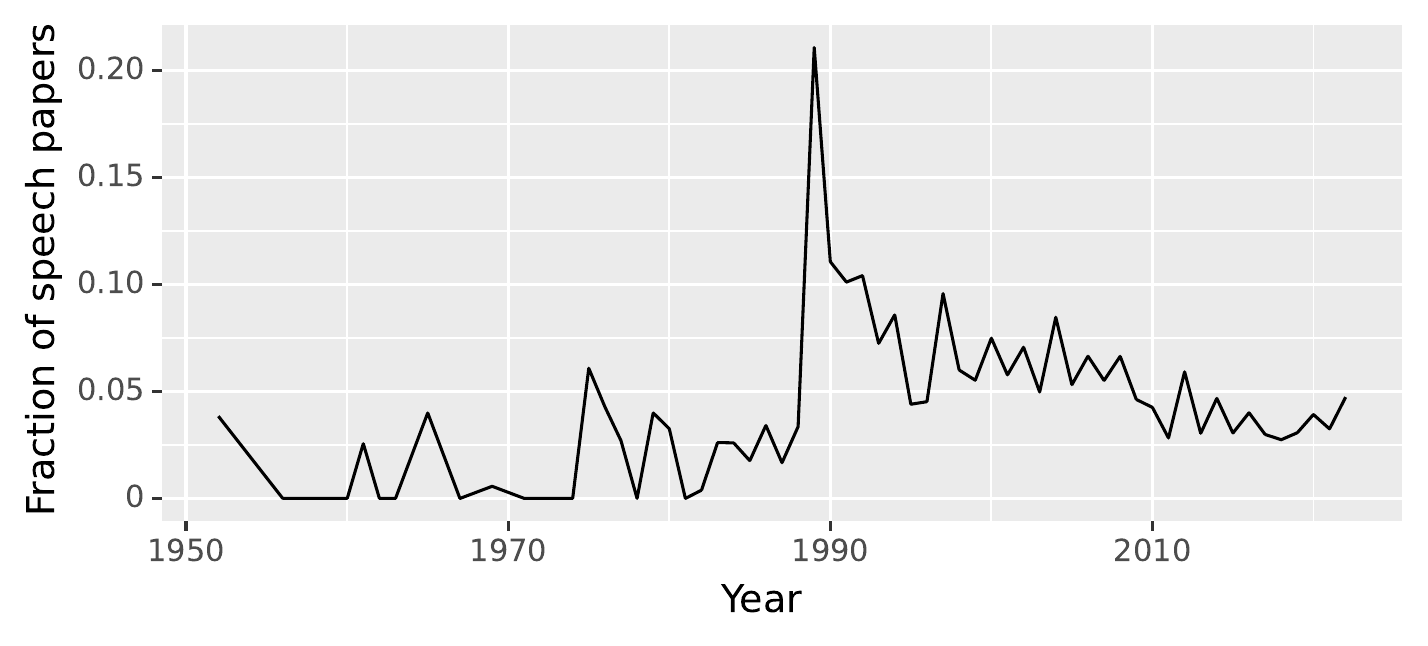}
  \caption{The proportion of papers in the ACL anthology up to year
    2022 which mention the words \textit{speech, spoken} or
    \textit{audio} in the title, excluding those with
    \textit{part(s)-of-speech} or \textit{speech act(s)}.}
    %The paper
    %database was downloaded from \url{https://aclanthology.org/} on
    %2023-01-09.}
  \label{fig:speech-fraction}
\end{figure}

\Cref{fig:speech-fraction} shows the proportion of NLP
papers explicitly mentioning speech-related terms in their title over
the years covered by the ACL anthology (1950 through 2022), which 
is a comprehensive database of NLP papers from a wide
variety of relevant conferences, workshops and journals.\footnote{\url{https://aclanthology.org/}}
%\footnote{The anthology does not include speech-specific venues such
%  as Interspeech organized by the Intenational Speech Communication
%  Association.}
The fraction of speech-focused NLP papers varies quite
a bit over the years, but mostly stays below 10\%. There is a large peak
going to 20\% in 1989, followed by three years with around 10\% of
speech papers.  A look at the underlying data reveals that the 1989
peak is associated with the inclusion in the anthology of the
proceedings of the Speech and Natural Language Workshop
\citep{hirshman1989overview} organized by the US Defense Advanced
Research Projects Agency (DARPA), and featuring 79 papers. This
workshop ran until 1992 and is thus largely responsible for the
four-year run of sizable representation of spoken language research
in the ACL anthology.

The overview of the last edition of this event
notes the then ongoing \textit{``paradigm shift in natural language processing
towards empirical, corpus based methods''}
\citep{marcus1992overview}. It is likely that this shift in NLP
methodology was at least partly driven by this workshop, the
associated DARPA program, and the resulting increased interaction
between researchers working on spoken and written language.

In recent years (since 2010) the proportion of NLP papers explicitly
mentioning spoken language has resolutely stayed below 6\%. While the
major ACL events typically include speech processing as a topic in
their calls for papers, as well as a track including the term
\textit{speech} in its name, such as \textit{Speech and
  Multimodality}, processing of spoken language it clearly a rather
minor concern of these conferences. Instead, speech work is
published in different venues organized by a separate speech
processing community.

\section{Spoken Language is Richer} 
\label{sec:harder}
While the primacy of the spoken modality as means of communication
is the consensus view in linguistics, \Cref{sec:unstated} identifies
unstated assumptions among NLP practitioners which amount to the
opposite view. Here I outline why these assumptions contradicting
the scientific view are not only incorrect but also detrimental to
progress on understanding and processing real human language.

\subsection{Key features of spoken language}
Speech and writing are two different modalities with different
affordances, and there is no straightforward mapping between them.
Some writing systems such as those used for
English, Arabic or Chinese do not even represent the phonology of the
spoken language in a direct way.
More crucially, writing only captures
a small proportion of the information carried in the equivalent audio
signal. Writing discards most of the information falling within the
general category of paralinguistic phenomena, such as that related to speaker identity,
speaker emotional state and attitude; likewise, information conveyed
by speech tempo and amplitude, including most of suprasegmental
phonology such as intonation and rhythm is typically not present in
writing.
In addition to the auditory
signal, oral spoken language can also feature visual clues in the form
of accompanying gestures, facial expressions and body posture. Sign
languages rely on the visual channel exclusively, and in fact there
are no widely used writing systems for any of them
\citep{grushkin2017writing}.  Unlike most text, speech 
also typically contains a variable amount of channel noise
\citep{shannon1948mathematical} such as environmental sounds.

Natural spontaneous speech contains fillers, hesitations, false
starts, repairs and other disfluencies \citep{dinkar2023fillers} which
are usually edited out in the written form of language. Even more
critically, spontaneous speech typically takes the form of a dialog
between two or more participants. Dialog is unlike common written
genres: crucially it features turn-taking behavior which is governed
by complex and incompletely understood rules
\citep{SKANTZE2021101178}. These features of natural dialog also mean
that the traditional cascaded approach of ASR followed by NLP faces
serious limitations, not least due to low ASR performance in this
regime \citep{szymanski-etal-2020-wer}, but also due to its inherently
interactive nature.

For all these reasons, spoken language is more informationally rich
than written language;\footnote{One exception to this general pattern
  is the presence of two spatial dimensions in written language, and
  the role of 2D layout in textual publications.} the same factors also make it more
variable, complex and noisy, and consequently more challenging for automated
processing \citep{shriberg05_interspeech}. 
Thus any understanding of language as a human faculty gained via the written modality does
not necessarily generalize to the
spoken modality. The same is also the case about language
applications: for example the successes and shortcomings of
state-of-the-art text chatbot systems
\citep[e.g.][]{NEURIPS2020_1f89885d} are likely to be substantially
different from those of spoken dialog systems.

\subsection{Challenges of speech}
As an illustrative example, let us consider the effectiveness of self-supervision:
inducing representations of words and phrases from just
listening to speech or reading text. For text, this general family of
methods has been successful since around the time of Latent Semantic Analysis \citep{dumais2004latent}, and 
 currently large written language models exhibit a
 constantly expanding range of abilities \citep{weiemergent}.
 In contrast, self-supervision with spoken language
has met with a limited amount of success only in the last few years
\citep[e.g.][]{baevski2020wav2vec,hubert}, and these models as of now are
usually only fine-tuned on the task of ASR. One obvious difference is
that items such as words and morphemes are either explicitly delimited or
easily discovered in text, but finding them is an unsolved research
problem in speech, due to the inherent variability of this modality.

On the other hand, learning spoken language becomes much more
tractable when self-supervision is augmented with
grounding in perception. The cross-modal
correlations, though unreliable and noisy, are often sufficient to
substantially facilitate the discovery and representation of words \citep{peng22c_interspeech,nikolaus-etal-2022-learning} and
syllables \citep{peng2023syllable} in
spoken language.
For written language, grounding in the visual modality has also been
found to help in some cases \citep[e.g.][]{tan-bansal-2020-vokenization}
but it does not appear crucial, as the dominance of text-only
language models demonstrates.

Since spoken language is richer in information content, it should in
principle be possible to exploit this extra signal for improving
performance. One obstacle to such developments is
the increased variability and channel noise. Perhaps less
obviously, a second obstacle is that widely used benchmarks are often
designed in a way which obstructs obtaining such gains. For
example the 2021 Zerospeech challenge
\citep{https://doi.org/10.48550/arxiv.2104.14700} which aimed to
benchmark spoken language modeling, evaluates systems according to the
following criteria: phoneme discrimination, word recognition,
syntactic acceptability and correlation to human judgments of
word similarities. None of these metrics would benefit much from
modeling speaker characteristics, speech tempo, pitch, loudness or
even suprasegmental phonology. Except for the first one,
these metrics would be very well suited for models trained exclusively
on written language. The combined effect of these two obstacles was
evident in the results of Zerospeech 2021 where written-language
toplines, such as RoBERTa
\citep{https://doi.org/10.48550/arxiv.1907.11692}, outperformed 
spoken language models on the latter three metrics, often by large
margins. 

\section{Unifying Speech Processing and NLP}
\label{sec:unify}

As evident from the examples highlighted above, spoken language is in
some ways quite different from written language and presents a
distinct set of challenges and potentials. In order to understand how
much progress the fields of speech and NLP are making in understanding
and implementing human language, we need to take speech
seriously \textit{qua} language, not just a cumbersome modality, and
measure our progress accordingly.

\subsection{Converging methodology}
The time is ripe for a closer integration of the speech and NLP
communities and for a unified computational science of language.  The
set of methodologies used in speech and text processing used to be
quite distinct in the past.  Since the adoption of deep learning
both fields have converged to a large extent: currently
the state-of-the-art models for both spoken and written language rely
on transformer architectures \citep{NIPS2017_3f5ee243} self-trained on
large amounts of minimally preprocessed data, with optional
fine-tuning. The technical communication barriers across disciplinary
boundaries are thus much lower. The recent emergence of the concept of
\emph{textless NLP} \citep{lakhotia-etal-2021-generative} exemplifies
the potential of unifying these two fields.

%Culture barriers have also subsided to some extent: even though the
%two communities publish in largely disjoint venues with quite
%different publication practices, both use preprint servers such as
%ArXiv and thus have easy access to each other's work.

\subsection{Opportunities}
The following paragraphs outline the most important benefits of making
NLP more natural, ranging from basic science to practical
applications.

\paragraph{Modeling language acquisition.}
An increased attention to spoken language within NLP has the potential
to lead to a more realistic understanding of how well our current
methods can replicate key human language abilities. Acquiring language
under constraints that human babies face is the big one. There
is a large amount of work on modeling human language acquisition
which uses exclusively written data (at best transcribed from the original
audio). Hopefully by this point the reader will be
convinced that the relevance of this work to the actual issue under
consideration is highly questionable. We stand a much better
chance of figuring out human language acquisition if we refocus
attention on spoken language. %The technology is there to facilitate
%this shift.

\paragraph{Data efficiency.}
\citet{linzen-2020-accelerate} argues convincingly for language models which are
human-like in their data-efficiency and generalization
capabilities. It is, however, unclear
whether these properties can even be properly evaluated via the medium
of written language. Since the informational density and the
signal-to-noise ratio in written vs spoken language are so very different, it
makes little sense to compare human children with
language models trained on text. 
Furthermore, the challenges of pure self-supervision may
motivate us to take seriously the impact of
grounding in perception and interaction, which humans use universally
as a learning signal.

\paragraph{Unwritten languages.}
Many modes of human communication lack standard
written representation. These range from major languages spoken by
millions of people such as Hokkien \citep{mair2003forget}, to small or non-standard language
varieties, to sign languages. Shifting the emphasis of NLP
research from text to the primary, natural oral and gestural
modalities will benefit the communities using these varieties.

\paragraph{Spoken dialog systems.}
\citet{dingemanse-liesenfeld-2022-text} argue that language technology
needs to transition from the text to talk, and provide a roadmap of
how to harness conversational corpora in diverse languages to effect
such a transition.  Indeed, one of the most obvious benefits of spoken
language NLP would be dialog systems that do not need to rely on ASR
and are able to exploit the extra information lost when transcribing
speech, enabling them to understand humans better and interact with
them in a more natural way.

\paragraph{Non-textual language data.}
Finally, there is a large and increasing stream of non-textual language
data such as podcasts, audio chat channels and video clips. Processing
such content could also benefit from an end-to-end holistic treatment
without the need of going through the lossy conversion to text.

\subsection{Recommendations}
If you are an NLP practitioner and view spoken language as
outside the scope of your field, reconsider. Getting into speech
processing does require understanding its specifics, but it is not as
technically daunting as it used to. Conversely, if you are a speech
researcher, consider that ASR and text-to-speech is not all there
is: we can get from sound to meaning and back without going through
the written word.
Both fields would do well to consider the
whole of human language as their purview. Increased 
collaboration would benefit both communities, and more importantly,
would give us a chance of making real progress towards
understanding and simulating natural language.

\section{Limitations}
The main limitation of this paper is the one applying to any opinion
piece: it is subjective and personal, as the views of the authors are
inherently limited by their expertise and experience. More
specifically, this paper argues for an increased interaction between
the speech and NLP communities, but the author is more strongly
embedded in the latter, and thus addresses this audience primarily.
Additionally, the short paper format imposes significant constraints
on the amount of nuance, detail and discussion of relevant literature,
and thus readers may find some of the claims to be less strongly
supported and less hedged than would be ideal, or proper in a longer
treatment of this topic.

\section*{Acknowledgements}
I would like to thank Hosein Mohebbi, Afra Alishahi, Mark Dingemanse,
Tanvi Dinkar, Piotr Szymański and three anonymous reviewers for their
valuable feedback on this paper.
\bibliography{biblio,anthology}
\bibliographystyle{acl_natbib}\balance
\end{document}